\documentclass[sigconf]{acmart}

\AtBeginDocument{%
  }

\copyrightyear{2026}
\acmYear{2026}
\setcopyright{cc}
\setcctype{by}
\acmConference[KDD 2026] {Proceedings of the 32nd ACM SIGKDD Conference on Knowledge Discovery and Data Mining V.2}{August 9--13, 2026}{Jeju Island, Republic of Korea.}
\acmBooktitle{Proceedings of the 32nd ACM SIGKDD Conference on Knowledge Discovery and Data Mining V.2 (KDD 2026), August 9--13, 2026, Jeju Island, Republic of Korea}
\acmISBN{979-8-4007-2259-2/2026/08}
\acmDOI{10.1145/3770855.3817573}
\usepackage{graphicx}
\usepackage{booktabs}
\usepackage{xcolor}
\usepackage{colortbl}
\usepackage{array}
\usepackage{multirow}
\usepackage{makecell}
\usepackage{longtable}
\usepackage{algorithm}
\usepackage{algorithmic}
\usepackage{enumitem}
\usepackage{balance}

\begin{document}

\title{PlatformBid: An Auto-Bidding Benchmark from a Unified Advertising Platform's Perspective}

\author{Shengtian Yang}
\authornote{Equal contribution. Work done when Shengtian was an intern at Kuaishou.}
\affiliation{%
  \institution{Southeast University}
  \city{Nanjing}
  \country{China}
  }
\email{yangshengtian@kuaishou.com}

\author{Yewen Li}
\authornotemark[1]
\affiliation{%
  \institution{Kuaishou Technology}
  \city{Beijing}
  \country{China}
  }
\email{liyewen@kuaishou.com}

\author{Peng Jiang}
\affiliation{%
  \institution{Kuaishou Technology}
  \city{Beijing}
  \country{China}
}
\email{jiangpeng07@kuaishou.com}

\author{Zhiyi Lyu}
\affiliation{%
  \institution{Nanyang Technological University}
  \city{Singapore}
  \country{Singapore}
  }
\email{zhiyi.lyu@ntu.edu.sg}

\author{Bo An}
\affiliation{%
  \institution{Nanyang Technological University}
  \city{Singapore}
  \country{Singapore}
  }
\email{boan@ntu.edu.sg}

\author{Peng Jiang}
\affiliation{%
  \institution{Kuaishou Technology}
  \city{Beijing}
  \country{China}
}
\email{jiangpeng11@kuaishou.com}

\author{Qingpeng Cai}
\authornotemark[2]
\affiliation{%
  \institution{Kuaishou Technology}
  \city{Beijing}
  \country{China}
}
\email{caiqingpeng@kuaishou.com}

\author{Lei Feng}
\authornote{Corresponding authors.}
\affiliation{%
  \institution{Southeast University}
  \city{Nanjing}
  \country{China}
  }
\email{fenglei@seu.edu.cn}

\renewcommand{\shortauthors}{Shengtian Yang et al.}

\begin{abstract}

%

Real-time bidding is central to computational advertising, comprising three elements: Supply Side Platform (SSP) selling ad impressions, Demand Side Platform (DSP) bidding for advertisers, and Ad Exchange conducting auctions between them.
Traditional auto-bidding algorithms focus solely on the DSP side, maximizing advertiser conversions by adjusting bids against competitors.
However, current big ad platforms, such as social media and e-commerce companies, now integrate SSP, DSP, and Ad Exchange functions internally.
From such ad platforms' perspective, the goal of the auto-bidding algorithms is not only to maximize the advertisers' conversions, but also the total revenue of the platform.
Given the lack of platform-centric evaluation frameworks and the pressing need to advance auto-bidding research, we propose \textbf{PlatformBid} - the first comprehensive benchmark designed from a unified ad platform's perspective.
To accurately reflect the real-world auto-bidding scenarios, we define three representative settings:
(1) homogeneous competition with identical algorithms across advertisers, (2) heterogeneous competition with diverse algorithmic strategies, and (3) promotional competition where some advertisers surge budgets for boosting sales during promotional events like \textit{Black Friday}.
We systematically evaluate a broad spectrum of existing auto-bidding methods across these settings, encompassing classical control methods, RL-based methods, and recent generative methods.
Besides these methods, we further propose a novel auto-bidding method based on flow-matching, termed \textbf{BidFlow}, which leverages the flow-matching method's expressive policy representation to effectively handle dynamic competitive environments.
Experimental results demonstrate the effectiveness of our proposed BidFlow on the new settings. 
Online experiments on Kuaishou further show a +0.68\% improvement in target cost, providing deployment evidence for the offline-online consistency of PlatformBid.\footnote{Code is available at https://github.com/YsTvT/PlatformBid.}
\end{abstract}




\maketitle

\begin{figure}
    \centering
    \includegraphics[width=0.95\linewidth, page=4]{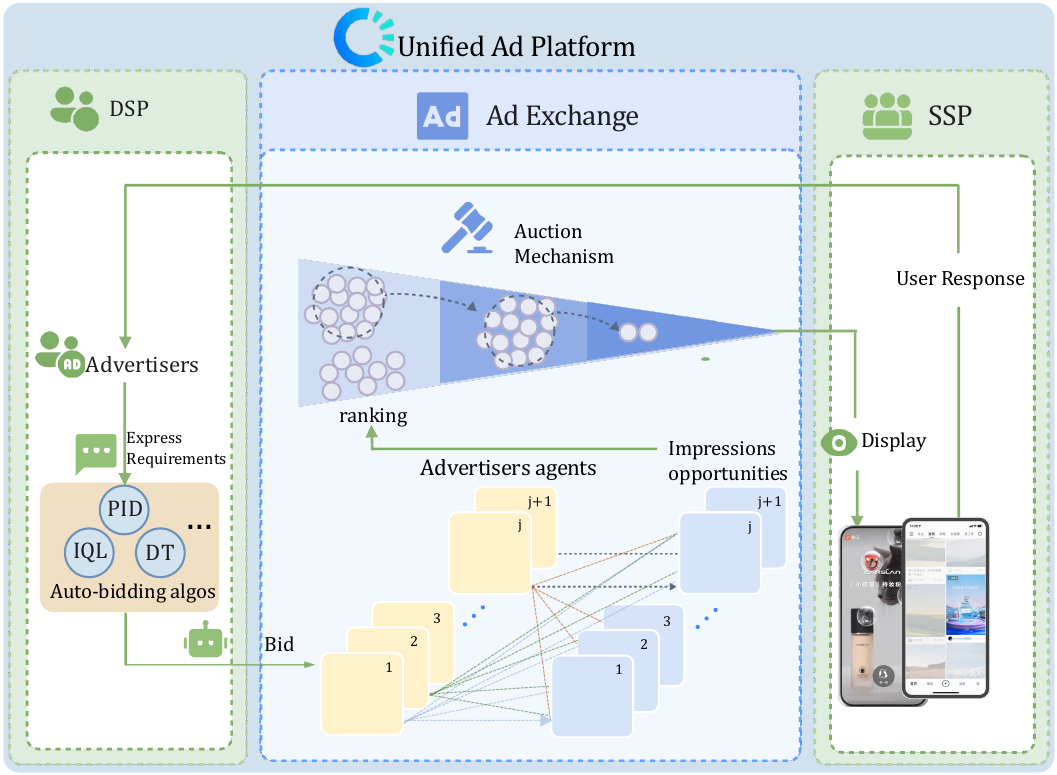}
    \caption{Overview of the unified ad platform. Previous benchmarks primarily focus only on the DSP side. However, as unified ad platforms integrate both DSP and SSP functionalities, a new benchmark from the platform perspective is urgently needed.}
    \label{fig:example}
\end{figure}

\begin{figure*}
    \centering
    \includegraphics[width=0.9\linewidth]{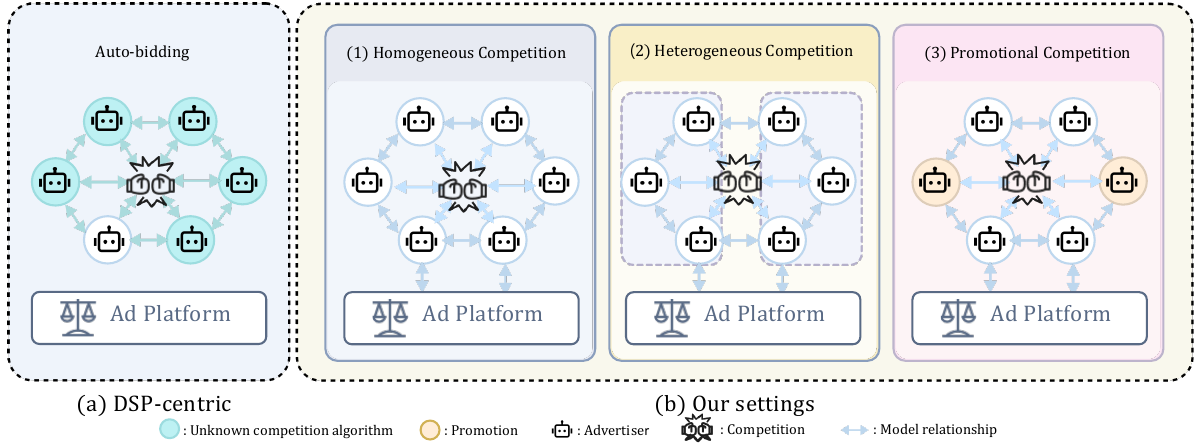}
    \vspace{-4mm}
    \caption{Comparison of PlatformBid settings with existing DSP-centric benchmarks. (a) DSP-centric benchmarks optimize only individual advertiser objectives, ignoring platform-level impacts. (b) Setting 1 (Homogeneous Competition): N advertisers with identical strategies compete dynamically. (c) Setting 2 (Heterogeneous Competition): N/2 baseline advertisers (white) compete against N/2 test advertisers (yellow) simultaneously. (d) Setting 3 (Promotional Competition): Selective budget amplification for specific advertisers (yellow) simulates promotional campaigns. Unlike DSP-centric approaches, PlatformBid jointly considers both advertiser and platform-level objectives.}
    \label{fig:benchmark}
\end{figure*}

\section{Introduction}

Real-time bidding (RTB) has emerged as a dominant paradigm in computational advertising~\cite{RTB}, significantly enhancing the efficiency of audience engagement and sales conversion on the Internet. As illustrated in Figure~\ref{fig:example}, the RTB process operates as follows: when an impression opportunity arises from 
a user's activity on a supply-side platform (SSP), such as watching videos on a media app, a bid request is broadcast to all competing advertisers via an Ad Exchange. Advertisers then leverage auto-bidding services~\cite{auto-bidding} provided by demand-side platforms (DSPs) to determine bid prices in real-time. 
The Ad Exchange then selects the highest bidder to display the ad, charging the winning advertiser the market prices, typically the second-highest bid price in a generalized second-price (GSP) auction setting~\cite{GSP}.
This paradigm greatly simplifies advertiser operations, as they only need to specify high-level objectives and constraints (\textit{e.g.}, target CPA, budget limits) to the DSP, making auto-bidding algorithms indispensable in modern computational advertising.
Existing auto-bidding research has predominantly adopted a DSP-centric perspective~\cite{yuan2014survey,wang2017display,zhang2014optimal,ou2023survey}, where DSP companies provide algorithmic bidding solutions designed to optimize individual advertiser objectives only. 
However, recent industry developments have fundamentally altered this landscape. The emergence and rapid growth of big media and e-commerce companies, such as Meta, TikTok, and Kuaishou, have driven substantial increases in intra-platform advertising activity. 
Therefore, these companies operate unified ad platforms that consolidate SSP, DSP, and Ad Exchange functionalities. 
This structural transformation demands a paradigm shift in the auto-bidding research: algorithms must now balance individual advertiser objectives~\cite{jin2018real,
multi-agentDB} with platform-level revenue optimization.

However, there is still a lack of evaluation benchmarks designed from a unified ad platform's perspective. Existing benchmarks in auto-bidding research have primarily focused on DSP-centric scenarios. A pioneering effort is the iPinYou dataset~\cite{iPinYou}, which provides real-world RTB logs from a DSP serving multiple advertisers across various ad exchanges. Building upon this paradigm, subsequent works have introduced benchmarks with richer features and larger scales. For instance, AuctionNet~\cite{su2024auctionnet} proposes a 
large-scale benchmark incorporating diverse advertiser types and competitive auction dynamics.
However, all these benchmarks are still designed for the DSP-centric auto-bidding research.
Given the pressing need to advance auto-bidding research, we propose the first comprehensive benchmark, termed \textbf{PlatformBid}, designed from a unified ad platform’s perspective.

To develop a benchmark that accurately and comprehensively captures the essence of platform-level auto-bidding scenarios, we formulate three representative settings in PlatformBid, as illustrated in Figure~\ref{fig:benchmark} (b):
(1) \textbf{Homogeneous competition}. This scenario simulates the platform-wide deployment of a new auto-bidding algorithm, where all advertisers adopt the same strategy. This necessitates evaluating both overall platform revenue and average advertiser performance. We model this through all advertisers employing identical algorithms.
(2) \textbf{Heterogeneous competition.} Platforms may ensemble multiple auto-bidding algorithms for different purposes, requiring assessment of both aggregate platform performance and inter-algorithm competitive dynamics. We simulate this by having half of the advertisers adopt a baseline method and the other half adopt an alternative method. This setting evaluates whether new algorithms can improve platform-level performance while avoiding adverse impacts on baseline advertisers.
(3) \textbf{Promotional competition.} Ad platforms derive substantial revenue from promotional events such as \textit{Black Friday}, during which certain advertisers significantly increase their budgets to achieve more conversions. We simulate this by selectively amplifying budgets for specific advertisers, examining auto-bidding performance under budget imbalance.

We systematically evaluate a broad spectrum of auto-bidding methods on PlatformBid across the three representative settings. 
The evaluated methods encompass: (1) Classical control methods, including PID controllers~\cite{chen2011real}; (2) Reinforcement learning (RL) methods, particularly offline RL approaches such as BCQ~\cite{fujimoto2019off} and IQL~\cite{kostrikov2021offline}; and (3) State-of-the-art generative methods, comprising Decision Transformer-based approaches~\cite{chen2021decision} such as GAS~\cite{li2024gas} and GAVE~\cite{gao2025gave}, as well as diffusion-based methods~\cite{diffuser1,diffuser2,generate3,generate4} including CBD~\cite{li2025cbd}.
Beyond these existing approaches, we propose \textbf{BidFlow}, a novel flow-matching-based auto-bidding method specifically designed for dynamic competitive environments. BidFlow leverages flow-matching's expressive policy representation to effectively capture the complex dynamics of ad auctions and employs Q-value-guided distillation to distill a multi-step flow model into an efficient one-step policy.
BidFlow achieves the best or competitive performance in most settings, while the heterogeneous sparse setting exposes a remaining challenge under extremely limited feedback. 
Furthermore, online experiment results on the Kuaishou ad platform, \textit{e.g.}, a {+0.68\%} increase of target cost, validate both BidFlow's effectiveness in more complex industrial scenarios and PlatformBid's offline-online consistency, confirming the benchmark's practical value for guiding real-world auto-bidding algorithm development.
In summary, the contributions of this work can be summarized as follows:
\begin{itemize}[leftmargin=0.5cm]
    \item We propose {PlatformBid}, the first comprehensive benchmark designed from a unified ad platform's perspective, formulating three representative settings—homogeneous competition, heterogeneous competition, and promotional competition.
    \item We evaluate comprehensive auto-bidding methods including classical control approaches, RL methods, and generative approaches, and propose BidFlow, a flow-matching-based method that achieves superior performance through expressive policy representation and Q-value-guided one-step distillation.
    \item We provide extensive analysis with diverse baselines and validate offline-online consistency through the online experiment, especially a significant {+0.68\%} increase of target cost, confirming PlatformBid's practical value for guiding real-world auto-bidding algorithm development.
\end{itemize}

\section{Related Work}
\noindent\textbf{Auto-bidding Benchmarks.} 
Systematic evaluation of auto-bidding algorithms has long relied on proprietary datasets~\cite{chen2011real,USCB,GSP}, limiting method comparison and reproducibility due to lack of unified standards.
The iPinYou dataset~\cite{iPinYou} pioneered public benchmarking with real-world RTB logs from a DSP serving multiple advertisers across ad exchanges. However, it primarily focuses on DSP-centric scenarios that optimize individual advertiser objectives. Subsequent benchmarks follow this paradigm while introducing richer features and larger scales—such as AuctionGym~\cite{auctiongym} and AdCraft~\cite{AdCraft}—but still optimize individual advertiser strategies in isolation.
AuctionNet~\cite{su2024auctionnet} advanced the field with the first large-scale public dataset and corresponding benchmark. However, its evaluation protocol sequentially replaces advertisers against fixed historical bids by replaying logged ad traffic, failing to capture the competitive dynamics where multiple advertisers simultaneously adjust strategies.
In contrast, as illustrated in Figure~\ref{fig:benchmark}, PlatformBid introduces dynamic multi-advertiser competition across three representative settings, enabling comprehensive assessment of both individual advertiser performance and collective platform revenues.

\noindent\textbf{Auto-bidding Methods.} 
Auto-bidding algorithms have evolved through three main paradigms, including classical control approaches, RL methods, and Generative methods.  
\textit{Classical control approaches}~\cite{chen2011real,non1,non2,non3,non4,non5,non6,non7}, such as PID controllers~\cite{chen2011real}, provide basic automation through feedback control mechanisms but lack adaptability to complex market dynamics.
\textit{Reinforcement learning methods}~\cite{rl1,rl2,li2026phgpo,yang2026pamoe} have significantly advanced the field. Online algorithms like USCB~\cite{USCB} and MAAB~\cite{multi-agentDB} optimize bidding strategies through direct environment interaction, while offline methods including BCQ~\cite{fujimoto2019off} and IQL~\cite{kostrikov2021offline} improve sample efficiency by learning from historical logs without risky online exploration.
\textit{Generative methods}~\cite{metadiffuser,d4rl,diffuserlite} offer powerful sequence modeling capabilities. Decision Transformer~\cite{chen2021decision,dt} formulates bidding as conditional sequence generation, avoiding value function estimation instability. GAS~\cite{li2024gas} extends this with post-training search using multiple critic networks, while GAVE~\cite{gao2025gave} introduces value-guided exploration to address the dataset quality limitation. Diffusion-based methods such as DiffBid~\cite{guo2024aigb} and CBD~\cite{li2025cbd} provide trajectory-level planning-based auto-bidding, with CBD achieving notable improvements in sparse-reward scenarios through its completer-aligner framework.
However, diffusion models suffer from slow inference due to iterative denoising, limiting deployment in latency-sensitive bidding systems. 
As these methods have only been validated in DSP-centric benchmarks, we find their performance in PlatformBid is limited due to the increased complexity of environment dynamics.
To address this challenge, we propose BidFlow, which leverages flow matching for efficient inference while incorporating Q-learning guidance for optimization.

\section{PlatformBid Benchmark}

In this section, we introduce our PlatformBid benchmark for platform-centric auto-bidding research. 
We first formulate the auto-bidding problem from a unified platform's perspective, highlighting key differences from traditional DSP-centric formulations. We then describe the benchmark's data preparation. Subsequently, we present three evaluation settings designed to reflect real-world auto-bidding scenarios. Finally, we detail the evaluation metrics that capture both advertiser-level and platform-level performance.

\subsection{Problem Statement}
Consider an online advertising scenario involving a sequence of \(H\) impression opportunities, \textit{i.e.}, impressions traffic, each identified by index \(i\), where PlatformBid simulates it by replaying a traffic log of impressions.
The advertiser secures an opportunity when the bid \(b_i\) exceeds competing bids, resulting in an associated cost \(c_i\).
PlatformBid employs a Generalized Second-Price (GSP) auction, thus $c_i$ equals the second-highest bid among all competing advertisers.
The optimization objective seeks to maximize aggregate value from all opportunities, formulated as \(\sum_i o_i v_i\), where \(v_i\) represents the value of opportunity \(i\) and \(o_i\) serves as a binary indicator of winning status. 
Beyond budget limitations, the system must satisfy various economic requirements, including constraints on unit costs for specific conversion events such as cost-per-action (CPA) \citep{USCB}.
Consequently, the auto-bidding optimization problem under constraints can be formulated as:
\begin{equation}
\begin{aligned}
\max \quad & \sum\nolimits_i o_i v_i \\
\text{s.t.} \quad & \sum\nolimits_i o_i c_i \leq B,\\
& \frac{\sum\nolimits_i o_i c_{ij}}{\sum\nolimits_i o_i p_{ij}} \leq C_j, \quad \forall j, \\
& \sum\nolimits_i e_{il}\leq O_l, \quad \forall l,\\
& o_i\in \{0, 1\}, \quad \forall i,
\label{eq:goal}
\end{aligned}
\end{equation}
where $B$ denotes the advertiser's total budget, $p_{ij}$ represents the performance indicator associated with the \(j\)-th advertiser-specified constraint with upper bound \(C_j\), $e_{il}$ quantifies the impact on the $l$-th platform-level performance metric, and $O_l$ specifies the corresponding platform objective threshold. 
For instance, $O_l$ may stipulate that the aggregate platform revenue across all $H$ opportunities must not fall below that obtained under a baseline bidding strategy, analogous to the requirement in online A/A testing.
{A critical distinction} from conventional DSP-centric formulations is the explicit incorporation of platform objectives $O_l$.
The subsequent sections examine three distinct problem settings that emerge from different specifications of $O_l$.

Prior research \citep{USCB, li2024gas, guo2024aigb} have predominantly adopted a simplified auto-bidding strategy formulated as:
\begin{equation}
b_i = \lambda_0 v_i + \sum\nolimits_{j=1}^{J} \lambda_j {p_{ij}}{C_j}, 
\label{eq:action_bid}
\end{equation}
where \(b_i^i\) denotes the final bid amount for opportunity \(i\). 
The coefficients \(\lambda_j\) constitute the bidding parameters. 
In real-world ad platforms, auto-bidding algorithms typically update these bidding parameters at relatively coarse time granularities---such as 30-minute intervals---rather than reacting to individual impression events occurring at millisecond scales.
This temporal characteristic provides sufficient computational headroom for sophisticated auto-bidding approaches to process parameter adjustment requests.

\subsection{Data Preparation}
The auto-bidding task can be formulated as a sequential decision-making task. 
At each timestep $t$ within a bidding period, the agent observes state $s_t\in\mathcal{S}$ that encodes the current advertising status, and subsequently selects action $a_t\in\mathcal{A}$ to determine bidding parameters.
The advertising environment evolves according to an unknown transition dynamic $\mathcal{T}$, where the next state is determined by both historical context and current observations.
Upon each state transition, the environment generates reward $r_t$, which quantifies the value accrued toward the campaign objective during period $t$.
This process repeats until the bidding horizon concludes (\textit{e.g.}, at the end of a day).
The key component data of the formulation are detailed below:
\begin{itemize}
    \item \textbf{State $s_t$}: A collection of campaign-level features describing the current advertising status, including temporal and budget-related information like remaining time, residual budget, expenditure rate, and KPI satisfaction ratios for each constraint.
    
    \item \textbf{Action $a_t$}: Adjustments to the bidding parameters \(\lambda_j\) for \(j = 0, \ldots, J\) at timestep \(t\), represented as \((a_t^{\lambda_0}, \ldots, a_t^{\lambda_J})\).
    
    \item \textbf{Reward $r_t$}: The conversion value accumulated during the interval from $t$ to $t+1$, serving as the optimization signal for the campaign objective.
\end{itemize}
For benchmarking learning-based auto-bidding methods, we adopt the mainstream offline learning paradigm, which trains policies on historical bidding logs rather than through live online interaction, thereby providing critical safety guarantees.

Our benchmark is compatible with any datasets containing the above essential MDP components, \textit{i.e.}, states, actions, and rewards.
Currently, PlatformBid is built upon the AuctionNet~\cite{su2024auctionnet} dataset, which contains 480K training trajectories collected from 48 advertisers operating under diverse budget and CPA constraints, along with impression traffic logs for testing simulation. The dataset provides two variants based on reward sparsity: Dense and Sparse.
While AuctionNet focuses on single-advertiser optimization, we extend it with a critical distinction: simultaneous multi-advertiser policy deployment within a shared auction environment.
To prevent advertisers from colluding to submit low bids that would harm the platform's total revenue, a floor price mechanism is adopted.
This architectural shift—from isolated single-agent evaluation to interactive multi-agent simulation—enables advertisers to respond to each other's bidding strategies in real time, thereby capturing competitive dynamics in the ad environment.
We further introduce platform-level evaluation metrics to assess platform-level performance beyond individual advertiser objectives.





\subsection{Evaluation Settings}
PlatformBid evaluates auto-bidding algorithms under realistic multi-advertiser competitive dynamics from a platform-centric perspective. A truly effective auto-bidding algorithm must satisfy three criteria: (1) strong performance under universal adoption, ensuring platform-wide efficiency; (2) robustness when competing against diverse strategies without crowding out other participants; and (3) generalization under non-stationary conditions such as promotional events. These criteria motivate our three evaluation settings.
\paragraph{\textbf{Setting 1: Homogeneous Competition.}}
The first setting, depicted in Figure~\ref{fig:benchmark}(b)-left, evaluates platform-wide performance when all $N$=48 advertisers deploy identical bidding policies. Each advertiser uses the same strategy and competes for impressions through the GSP auction mechanism. Performance is measured by averaging scores across all advertisers, capturing aggregate platform outcomes rather than individual advertiser success.

This setting directly mirrors the online experiments where a new auto-bidding algorithm is simultaneously deployed to all advertisers.
In such deployments, the platform must evaluate whether the new algorithm improves performance at both the advertiser level and platform level compared to the in-production baseline method. 
Under homogeneous strategy adoption, emergent competitive dynamics reveal whether the algorithm achieves stable market equilibrium or triggers systemic inefficiencies such as bid inflation or collusive behavior.
This homogeneous competition scenario thus provides a controlled testbed for evaluating platform-wide algorithm launch, measuring both aggregate revenue generation and average advertiser welfare under identical strategy adoption.

\paragraph{\textbf{Setting 2: Heterogeneous Competition.}} The second setting, illustrated in Figure~\ref{fig:benchmark}(b)-middle, evaluates robustness under strategic diversity by partitioning advertisers into two groups: $N/2=24$ baseline advertisers running a fixed reference policy (vanilla Decision Transformer~\cite{chen2021decision}, a widely adopted method in industry) and $N/2=24$ test advertisers running the algorithm under evaluation. This setting captures the deployment reality where new algorithms are rolled out to a subset of advertisers while others continue using existing solutions.

This setting directly mirrors the online test with an ensemble of strategies. In practice, platforms may not deploy a single algorithm to all advertisers simultaneously, as different scenarios have unique requirements. Instead, new algorithms are typically tested on a subset of target advertisers while others continue using existing solutions. This approach enables controlled performance comparison under real competitive dynamics.
Only algorithms demonstrating gains for test advertisers without degrading baseline advertiser outcomes satisfy the requirements for production deployment.

\paragraph{\textbf{Setting 3: Promotional Competition.}} The third setting, shown in Figure~\ref{fig:benchmark}(b)-right, evaluates generalization under non-stationary conditions by simulating promotional events. A subset of advertisers (7 out of 48 advertisers in this benchmark) receive doubled budgets, mimicking the budget surges accompanying sales promotions. All advertisers deploy the same algorithm, with metrics reported separately for promotional advertisers, non-promotional advertisers, and platform-wide aggregates.

This setting simulates promotional campaigns that constitute a major revenue source for ad platforms, such as \textit{Black Friday}, where participating advertisers dramatically increase budgets to boost sales. The setting tests two critical capabilities: promotional advertisers must effectively scale their bidding strategies to utilize expanded budgets without violating CPA constraints, as optimal behavior at doubled budget diverges significantly from training distributions; meanwhile, non-promotional advertisers must maintain competitive performance despite intensified competition from budget-rich rivals. This validates whether algorithms can adapt to the non-stationary market dynamics characteristic of major promotional events while ensuring fairness across advertisers with different economic resources.

\subsection{Evaluation Metrics}
Our evaluation framework employs comprehensive metrics widely adopted in industrial advertising platforms, encompassing measurements at both platform and advertiser levels. 

Platform-centric metrics measure aggregate system outcomes:
\begin{itemize}
    \item \textbf{Conversion}: Average conversion value generated across all advertisers.
    \item \textbf{Budget Utilization}: Average fraction of allocated budgets spent, indicating market liquidity and inventory efficiency.
\end{itemize}
Advertiser-centric metrics assess individual advertiser experience and constraint satisfaction:
\begin{itemize}
    \item \textbf{CPA Ratio}: Average ratio of realized CPA to target CPA across advertisers, where lower values indicate better ROI.
    \item \textbf{CPA Ratio Variance}: Variance of CPA ratios, measuring consistency of constraint satisfaction across advertisers.
    \item \textbf{Exceed Rate}: Fraction of advertisers violating CPA constraints (realized CPA exceeding target CPA).
    \item \textbf{Qualified Rate}: Fraction of advertisers achieving CPA ratios within the acceptable range [0.8, 1.2].
\end{itemize}
A comprehensive metric balances platform revenue with advertiser satisfaction could be the \textbf{Score}, calculated as
\begin{equation}
\text{Score}_i = \begin{cases}
\text{Conversion}_i & \text{if } \text{CPA}_i \leq \text{Target}_i \\
\text{Conversion}_i \times \left(\frac{\text{Target}_i}{\text{CPA}_i}\right)^2 & \text{otherwise}
\end{cases}
\end{equation}
This formulation reflects the platform's preference for sustainable growth—high conversions achieved through constraint violations are penalized, as such patterns erode advertiser trust and long-term platform health.

\noindent More \textbf{benchmark implementation details} are in Appendix \ref{app:benchmark_imple}.

\begin{figure}[t!]
    \centering
    \includegraphics[width=1\linewidth, page=7]{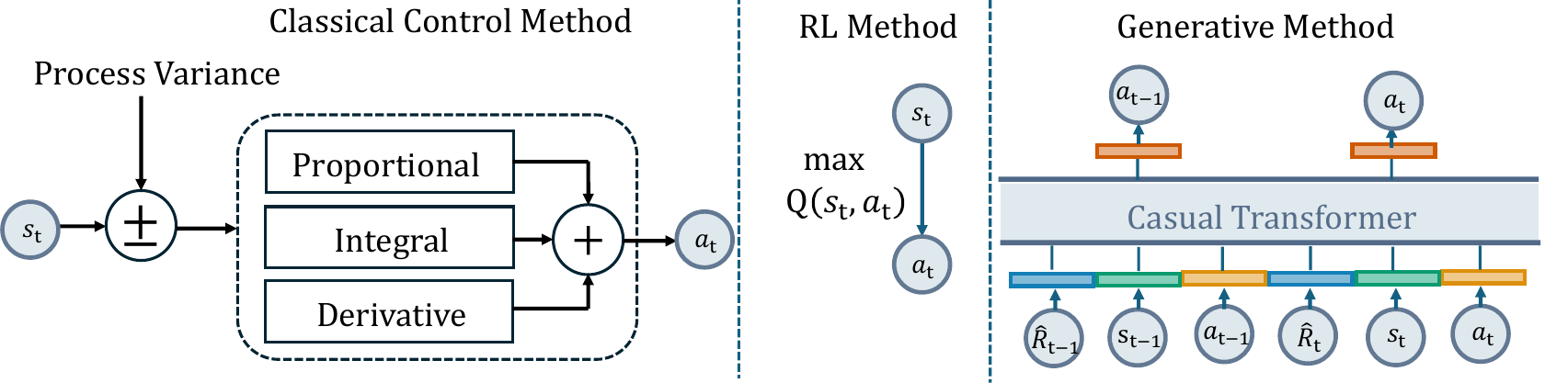}
    \caption{Three categories of auto-bidding algorithms.}
    \label{fig:figure3}
\end{figure}

\begin{figure*}[t!]
    \centering
    \includegraphics[width=0.9\linewidth, page=4]{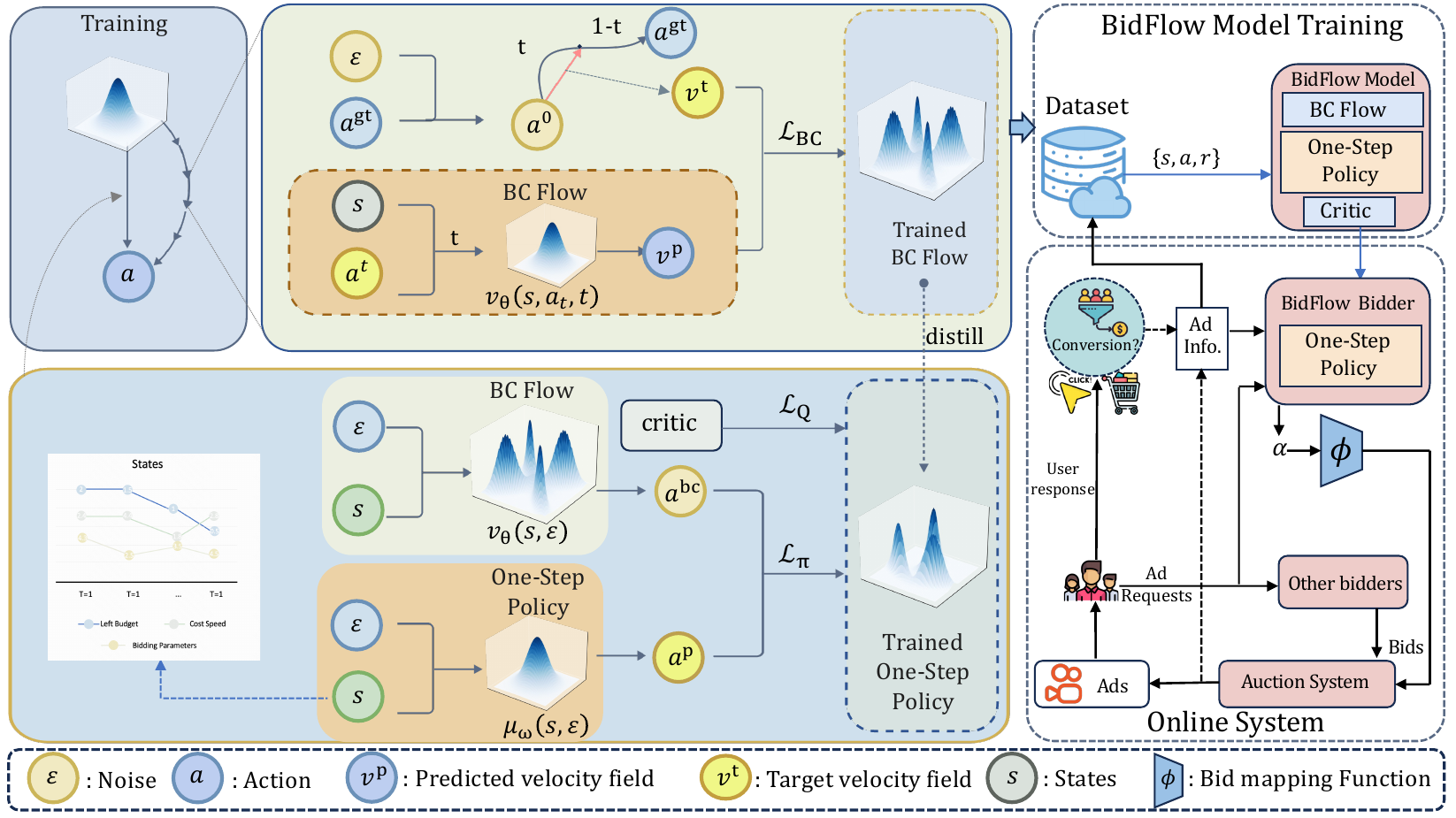}
    \caption{Overview of the BidFlow framework.
    Training (left): The BC flow policy learns a velocity field to transform noise into bids via flow matching. The one-step policy is trained to distill the BC flow policy while maximizing Q-values. Inference (right): Only the one-step policy is used, enabling efficient single-pass action generation without iterative sampling.}
    \label{fig:figure4}
\end{figure*}

\section{Methods}


\subsection{Auto-bidding Baselines}

As illustrated in Figure~\ref{fig:figure3}, existing auto-bidding methods can be categorized into three paradigms based on their inference mechanisms. We select representative methods from each for benchmarking:

\textbf{(1) Classical Control Methods} employ feedback control mechanisms without learning from data. We select {PID} controllers~\cite{chen2011real}, which compute bidding actions by combining proportional, integral, and derivative terms of the error between actual and target CPA. Given state $s_t$ and process variance, the controller outputs action $a_t$ through a fixed mathematical formula.

\textbf{(2) Reinforcement Learning Methods} learn a policy $\pi(a|s)$ that directly maps states to actions of maximum accumulative value. While RL methods can capture complex state-action relationships in a single forward pass, they typically assume unimodal action distributions (\textit{e.g.}, Gaussian policies) and the MDP assumption, limiting their ability to represent multi-modal bidding strategies required in dynamic auction environments. Following the offline learning paradigm, we include BCQ~\cite{fujimoto2019off} and IQL~\cite{kostrikov2021offline}, representing conservative and implicit Q-learning approaches, respectively.

\textbf{(3) Generative Methods} represent the current state-of-the-art in auto-bidding by formulating the problem as conditional generation. These methods fall into two sub-paradigms:

\textit{Decision-Transformer-based approaches} capture the joint distribution of return-to-go $R$, state $s$, and action $a$ through causal transformer architectures, autoregressively generating actions conditioned on target returns and historical sequences. We testify Decision Transformer (DT)~\cite{chen2021decision} in both standard and score-conditioned (DT score) variants, GAS~\cite{li2024gas} with post-training critic search, and GAVE~\cite{gao2025gave} with value-guided exploration. 
However, these methods typically model action distributions as deterministic or unimodal Gaussian, limiting their expressiveness.

\textit{Diffusion-based approaches} leverage the distributional modeling capacity of diffusion models to generate future states, then project them to corresponding actions through inverse dynamics models. We include CBD~\cite{li2025cbd} with its completer-aligner framework.

\textbf{More implementation details can be seen in Appendix~\ref{implementary details}.}

\begin{table*}[t!]
\centering
\caption{Comparison under PlatformBid setting 1: Homogeneous Competition. Arrow direction ``$\uparrow$'' and ``$\downarrow$'' denote the performance improvement direction. Best results are bolded.}
\label{tab:dense_vs_sparse}
\scriptsize
\renewcommand{\arraystretch}{0.1}
\resizebox{1.00\textwidth}{!}{
\setlength{\tabcolsep}{2.5mm}{
\begin{tabular}{@{}llccccccccc@{}}
\toprule
\textbf{Dataset} & \textbf{Metric} 
& \textbf{PID} & \textbf{IQL} & \textbf{BCQ} & \textbf{DT} & \textbf{DT score} & \textbf{GAS} & \textbf{GAVE} 
& \textbf{CBD} & \textbf{BidFlow} \\
\midrule
& Conversion $\uparrow$        & 183.98 & 331.79 & 236.48 & 347.79 & 323.06 & 353.00 & 334.52 & 293.98 & \textbf{358.90} \\
& Budget Util $\uparrow$   & 0.54 & 0.90 & 0.84 & 0.96 & 0.83 & 0.93 & \textbf{1.00} & 0.98 & 0.88 \\
& CPA Ratio $\downarrow$   & 1.39 & 1.00 & 1.24 & 1.04 & 0.91 & 0.97 & 1.14 & 1.28 & \textbf{0.88} \\
\textbf{Dense} & CPA Ratio Var $\downarrow$ & 0.69 & 0.07 & 0.06 & 0.06 & 0.03 & 0.03 & 0.12 & 0.16 & \textbf{0.03} \\
& Exceed Rate $\downarrow$ & 0.60 & 0.48 & 0.81 & 0.56 & 0.31 & 0.42 & 0.67 & 0.75 & \textbf{0.25} \\
& Qualified Rate $\uparrow$ & 0.27 & 0.60 & 0.42 & 0.60 & \textbf{0.73} & 0.67 & 0.44 & 0.38 & 0.67 \\
& Score $\uparrow$         & 151.74 & 289.74 & 169.60 & 287.62 & 311.77 & 314.27 & 241.57 & 189.34 & \textbf{348.04} \\
\midrule
& Conversion $\uparrow$        & 11.21 & 28.56 & 21.56 & 15.65 & 27.83 & 30.15 & 21.56 & 32.98 & \textbf{33.06} \\
& Budget Util $\uparrow$   & 0.62 & 0.87 & 0.44 & 0.45 & 0.58 & 0.65 & \textbf{1.00} & 0.82 & 0.78 \\
& CPA Ratio $\downarrow$   & 1.94 & 1.01 & 0.87 & 1.35 & 0.81 & \textbf{0.80} & 1.11 & 0.97 & 0.85 \\
\textbf{Sparse} & CPA Ratio Var $\downarrow$ & 0.62 & \textbf{0.09} & 0.32 & 0.89 & 0.15 & 0.10 & \textbf{0.09} & 0.16 & 0.11 \\
& Exceed Rate $\downarrow$ & 0.88 & 0.42 & 0.23 & 0.46 & \textbf{0.19} & 0.27 & 0.50 & 0.44 & 0.25 \\
& Qualified Rate $\uparrow$ & 0.15 & 0.38 & 0.31 & 0.21 & 0.27 & 0.31 & \textbf{0.46} & 0.40 & 0.31 \\
& Score $\uparrow$         & 6.82 & 22.76 & 20.98 & 13.72 & 26.62 & 28.45 & 15.52 & 28.27 & \textbf{30.59} \\
\bottomrule
\end{tabular}
}
}
\end{table*}

\subsection{BidFlow: A Flow-matching-based Auto-bidding Method}
However, we observe that existing auto-bidding baselines exhibit limited performance in these new settings. This degradation stems from the increasingly critical challenge of modeling complex, multi-modal bidding distributions—a challenge amplified by more dynamic market fluctuations and heightened competitive complexity at the platform level.
To address this, we propose advancing auto-bidding methods through flow matching, a state-of-the-art generative modeling paradigm that has demonstrated remarkable success in capturing complex data distributions across multiple domains~\cite{flow_matching, FQL, meanflow}.
Therefore, we propose a novel flow-matching-based auto-bidding method, termed BidFlow. 

As shown in Figure~\ref{fig:figure4}, BidFlow follows the typical flow-matching framework for offline RL~\cite{FQL}, comprising three key components: {(1)} a critic network $Q_\phi(s, a)$ that estimates state-action values; {(2)} a behavioral cloning (BC) flow policy $\mu_\theta(s, \varepsilon)$ trained via flow matching to capture the behavioral distribution in the offline dataset; and {(3)} a one-step policy $\mu_\omega(s, \varepsilon)$ that maximizes Q-values while being regularized through distillation from the BC flow policy. This one-step policy enables faster inference compared to the multi-step BC flow while achieving superior performance through Q-value guidance.

The critic is trained via standard temporal difference learning. Given transitions $(s, a, r, s')$ sampled from the offline dataset $\mathcal{D}$, the critic minimizes:
\begin{equation}
\mathcal{L}_Q(\phi) = \mathbb{E}_{(s,a,r,s') \sim \mathcal{D}} \left[ \left( Q_\phi(s, a) - r - \gamma \bar{Q}(s', \mu_\omega(s', \varepsilon')) \right)^2 \right],
\end{equation}
where $\bar{Q}$ denotes the target network updated via exponential moving average, and next actions are sampled from the one-step policy with $\varepsilon' \sim \mathcal{N}(0, I)$.

The BC flow policy learns a velocity field $v_\theta(t, s, a^t)$ that transforms Gaussian noise into actions through an ODE. It is trained exclusively with the flow matching objective:
\begin{equation}
\mathcal{L}_{\text{BC}}(\theta) = \mathbb{E}_{\substack{s,a \sim \mathcal{D}, a^0 \sim \mathcal{N}(0, I), \\ t \sim \text{Unif}([0,1])}} \left[ \|v_\theta(t, s, a^t) - (a - a^0)\|_2^2 \right],
\end{equation}
where $a^t = (1-t)a^0 + ta$ is the linear interpolation. At inference, actions are generated by numerically solving the ODE via the Euler method over $M$ steps.

The one-step policy $\mu_\omega(s, \varepsilon)$ learns to directly map noise to actions in a single forward pass. It is trained with a combined objective that balances behavioral regularization and value maximization:
\begin{equation}
\nonumber
\mathcal{L}_\pi(\omega) = \mathbb{E}_{s, \varepsilon} \left[ \|\mu_\omega(s, \varepsilon) - \mu_\theta(s, \varepsilon)\|_2^2 \right] + \alpha \mathbb{E}_{s, \varepsilon} \left[ -Q_\phi(s, \mu_\omega(s, \varepsilon)) \right],
\end{equation}
where the first term distills knowledge from the BC flow policy, the second term maximizes Q-values for outperforming the BC Flow, and $\alpha$ controls the regularization strength. 

All three components are trained jointly at each iteration. The complete procedure is summarized in Algorithm~\ref{alg:flowql}. 
At deployment, only the one-step policy $\mu_\omega$ is used for action selection, enabling efficient inference without iterative sampling.

\begin{algorithm}
\caption{Training Procedure of BidFlow}
\label{alg:flowql}
\begin{algorithmic}[1]
\REQUIRE Offline dataset $\mathcal{D}$, BC coefficient $\alpha$
\STATE Initialize critic $Q_\phi$, target critic $\bar{Q}_{\bar{\phi}}$, BC flow policy $v_\theta$, one-step policy $\mu_\omega$
\WHILE{not converged}
    \STATE Sample batch from $\mathcal{D}$
    \STATE Update $Q_\phi$ by minimizing TD loss (Eq. 4)
    \STATE Update $v_\theta$ by minimizing flow matching loss (Eq. 5)
    \STATE Compute distillation target $\mu_\theta(s, \varepsilon)$ via Euler method
    \STATE Update $\mu_\omega$ by minimizing $\mathcal{L}_\pi(\omega)$ 
    \STATE Soft update: $\bar{\phi} \leftarrow \tau \phi + (1-\tau)\bar{\phi}$
\ENDWHILE
\RETURN One-step policy $\mu_\omega$
\end{algorithmic}
\end{algorithm}

\begin{table*}[h]
\centering
\caption{Key metrics comparison under PlatformBid setting 2: Heterogeneous Competition. Each result pair (Target, Average) shows the average performance of only the target advertisers by the evaluated method and the average performance of all advertisers. Result pairs are bolded if ``Average'' is the best.
}
\vspace{-3mm}
\label{tab:setting2}
\scriptsize
\renewcommand{\arraystretch}{1.0}
\resizebox{1.00\textwidth}{!}{
\setlength{\tabcolsep}{1mm}{
\begin{tabular}{@{}llcccccccc@{}}
\toprule
\textbf{Dataset} & \textbf{Metric} 
& \textbf{PID} & \textbf{IQL} & \textbf{BCQ} & \textbf{DT score} & \textbf{GAS} & \textbf{GAVE} 
& \textbf{CBD} 
& \textbf{BidFlow} \\
\midrule
\multirow{3}{*}{\textbf{Dense}} 
& Conversion $\uparrow$        & 170.54, 254.17 & 309.17, 331.55 & 231.04, 265.90 & 333.17, 345.44 & 340.38, 347.55 & 355.71, 330.07 & 313.29, 303.46 & \textbf{346.88, 347.96} \\
& CPA Ratio $\downarrow$   & 2.78, 1.91 & 0.99, 1.03 & 1.19, 1.19 & 0.85, 0.96 & 0.91, 0.97 & 0.99, 1.06 & 1.20, 1.13 & \textbf{0.93, 0.93} \\
& Score $\uparrow$         & 143.21, 216.74 & 285.24, 293.86 & 165.21, 190.03 & \textbf{321.96, 321.92} & 315.81, 315.93 & 287.78, 262.78 & 229.86, 233.34 & 312.69, 308.83 \\
\midrule
\multirow{3}{*}{\textbf{Sparse}} 
& Conversion $\uparrow$        & 11.88, 21.65 & 38.46, 32.59 & 37.92, 34.42 & \textbf{44.38, 37.50} & 36.96, 33.86 & 29.30, 32.32 & 42.71, 31.84 & 38.08, 28.77 \\
& CPA Ratio $\downarrow$   & 2.08, 1.28 & 0.94, 0.77 & 0.63, 0.69 & \textbf{0.63, 0.65} & 0.72, 0.70 & 1.26, 0.84 & 0.89, 0.77 & 0.83, 1.13 \\
& Score $\uparrow$         & 9.52, 20.47 & 34.42, 30.31 & 37.78, 33.59 & \textbf{43.91, 36.26} & 36.47, 32.77 & 21.20, 28.29 & 38.71, 29.60 & 35.97, 26.51 \\
\bottomrule
\end{tabular}
}
}
\end{table*}

\begin{table*}[h]
\centering
\caption{Key metric comparison under PlatformBid Setting 3: Promotional Competition. ``Pro'' denotes promotional advertisers, ``Non'' denotes non-promotional advertisers, and ``All'' denotes the average performance across all advertisers.}
\vspace{-3mm}
\label{tab:setting3}
\scriptsize
\renewcommand{\arraystretch}{0.1}
\resizebox{\textwidth}{!}{
\setlength{\tabcolsep}{3mm}{
\begin{tabular}{@{}lll*{9}{c}@{}}
\toprule
\textbf{Dataset} & \textbf{Type} & \textbf{Metric}
& \textbf{PID} & \textbf{IQL} & \textbf{BCQ} & \textbf{DT} & \textbf{DT score} & \textbf{GAS} & \textbf{GAVE}
& \textbf{CBD}
& \textbf{BidFlow} \\
\midrule

& &Conversion $\uparrow$  & 550.71 & 674.00 & 543.14 & 673.14 & 682.57 & 690.43 & 657.71 & 456.86 & \textbf{727.29} \\
&Pro & CPA Ratio $\downarrow$ & 1.49 & 1.10 & 1.15 & 1.06 & 0.98 & 0.95 & 1.15 & 1.32 & \textbf{0.91} \\
& & Score $\uparrow$  & 519.80 & 587.70 & 429.46 & 555.49 & 634.28 & 661.49 & 456.13 & 347.37 & \textbf{701.48} \\
\cmidrule(lr){2-12}

& & Conversion $\uparrow$  & 124.93 & 294.10 & 234.17 & 307.85 & 316.43 & \textbf{319.95} & 304.32 & 252.17 & 310.61 \\
\textbf{Dense} &Non & CPA Ratio $\downarrow$ & 1.71 & 1.02 & 1.20 & 1.12 & 1.01 & 0.95 & 1.20 & 1.37 & \textbf{0.89} \\
& & Score $\uparrow$  & 111.46 & 247.66 & 171.42 & 230.79 & 271.85 & 277.90 & 203.71 & 169.46 & \textbf{297.25} \\
\cmidrule(lr){2-12}

& & Conversion $\uparrow$  & 187.02 & 349.50 & 279.23 & 361.12 & 371.29 & \textbf{373.98} & 355.85 & 282.02 & 371.38 \\
&All & CPA Ratio$\downarrow$ & 1.68 & 1.03 & 1.19 & 1.11 & 1.01 & 0.95 & 1.20 & 1.36 & \textbf{0.90} \\
& & Score $\uparrow$  & 171.01 & 297.25 & 209.05 & 278.14 & 321.73 & 333.84 & 240.52 & 195.41 & \textbf{356.20} \\
\midrule

& & Conversion $\uparrow$  & 24.86 & 71.00 & 54.57 & 26.00 & 64.71 & \textbf{78.86} & 48.57 & 60.86 & 71.86 \\
&Pro & CPA Ratio $\downarrow$ & \textbf{0.47} & 0.95 & 0.67 & 0.96 & 0.65 & 0.71 & 1.49 & 0.75 & 0.68 \\
& & Score $\uparrow$  & 24.85 & 64.24 & 54.57 & 25.38 & 64.71 & \textbf{78.85} & 24.70 & 59.57 & 71.86 \\
\cmidrule(lr){2-12}

& & Conversion $\uparrow$  & 5.95 & 27.44 & 17.49 & 15.05 & 23.90 & 23.93 & 22.93 & \textbf{29.27} & 28.80 \\
\textbf{Sparse} &Non & CPA Ratio $\downarrow$ & 1.37 & 0.98 & \textbf{0.81} & 1.12 & 0.86 & 1.14 & 1.64 & 0.99 & 0.92 \\
& & Score $\uparrow$  & 5.86 & 22.50 & 16.93 & 13.00 & 23.10 & 21.08 & 11.79 & 24.58 & \textbf{25.67} \\
\cmidrule(lr){2-12}

& & Conversion $\uparrow$  & 8.71 & 33.79 & 22.90 & 16.65 & 29.85 & 31.94 & 26.67 & 33.88 & \textbf{35.08} \\
& All & CPA Ratio $\downarrow$ & 1.24 & 0.98 & \textbf{0.79} & 1.10 & 0.83 & 1.08 & 1.62 & 0.95 & 0.88 \\
& & Score $\uparrow$  & 8.63 & 28.59 & 22.42 & 14.81 & 29.17 & 29.51 & 13.67 & 29.68 & \textbf{32.41} \\
\bottomrule
\end{tabular}
}
}
\end{table*}

\vspace{-5mm}
\section{Experiments}
\label{exp}
\subsection{Experimental Setup}
We evaluate representative methods spanning the classical control approach, offline RL, and generative approaches. PID~\cite{chen2011real} serves as the classical control baseline. For offline RL, we include BCQ~\cite{fujimoto2019off} and IQL~\cite{kostrikov2021offline}, which represent conservative and implicit Q-learning paradigms respectively. The generative model family includes Decision Transformer (DT)~\cite{chen2021decision} in both standard and reward-conditioned (DT score) variants, GAS~\cite{li2024gas} with post-training critic search, GAVE \cite{gao2025gave} with value-guided exploration, and CBD~\cite{li2025cbd} with its completer-aligner framework. 
All methods are trained on the AuctionNet datasets and evaluated on both dense and sparse variants of impression traffic test data.
Implementation follows the original papers~\cite{chen2011real,chen2021decision,li2024gas,gao2025gave,kostrikov2021offline,fujimoto2019off,li2025cbd,FQL} with hyperparameters tuned on a held-out validation set. For BidFlow, the BC flow policy uses $M=10$ Euler steps during training for distillation target computation. 
All results are averaged over 5 random seeds.
More details about implementation are shown in Appendix~\ref{implementary details}.
\subsection{Benchmark Implementation Details}\label{app:benchmark_imple}

PlatformBid is implemented as an extension to the AuctionNet simulation environment. The simulator processes impression opportunities chronologically, soliciting bids from all active advertisers and resolving auctions according to GSP rules with floor price enforcement. 

To prevent collusive bid suppression, which is a phenomenon observed particularly in sparse-reward environments where advertisers may collectively lower bids to reduce costs. To tackle this issue, we introduce a floor price mechanism. The floor price is set at 80\% of the original least winning cost from the AuctionNet dataset. If a winning bid falls below the floor price, the impression is not awarded and no transaction occurs. This mechanism preserves market integrity by ensuring that strategic bid reduction cannot extract value below a reasonable threshold, aligning individual incentives with platform revenues.

\begin{figure*}[t!]
    \centering
    \includegraphics[width=0.9\linewidth, page=7]{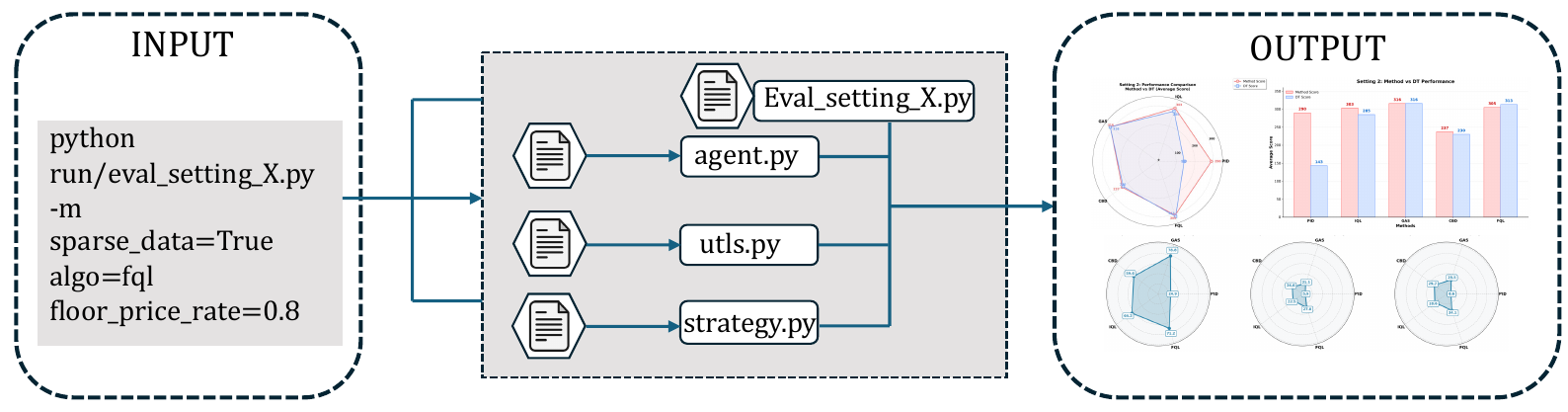}
    \caption{PlatformBid execution diagram. Users run evaluations as configurable experiments, where each setting loads its components from the respective Python modules.}
    \label{fig:5}
\end{figure*}

PlatformBid tackles its modularity goals via defining unifying abstractions over the auto-bidding evaluation pipeline. As depicted in Figure~\ref{fig:5}, components including agent strategies (\texttt{agent.py}), utility functions (\texttt{utls.py}), and bidding strategies (\texttt{strategy.py}) are gathered into evaluation scripts that are agnostic of the specific auto-bidding algorithm implementations. Structured command-line configurations allow users to easily specify the evaluation setting, dataset variant, competing algorithms, and auction parameters (\textit{e.g.}, floor price rate), making it possible to go directly from one-line inputs to comprehensive evaluation outputs.

\subsection{Result Analysis}

Tables~\ref{tab:dense_vs_sparse},~\ref{tab:setting2}, and~\ref{tab:setting3} present the results under key metrics across our three evaluation settings.  

{\textbf{\textit{BidFlow outperforms baselines.}}}
BidFlow demonstrates superior performance in Setting 1, attaining the highest scores across both dense and sparse conversion configurations while maintaining the lowest CPA exceed rate, indicating effective alignment between advertiser objectives and platform welfare. In Setting 3, BidFlow exhibits advantages for both promotional advertisers and non-promotional advertisers, demonstrating robust generalization under extreme budget imbalance conditions.
Nevertheless, our analysis also identifies specific limitations. Under Setting 2 with sparse rewards, BidFlow demonstrates performance degradation, with the baseline DT method struggling to acquire sufficient conversions. 
These findings provide important insights for developing auto-bidding methods in Setting 2 by explicitly modeling the influence between various auto-bidding methods, highlighting a promising direction for future research.


{\textbf{\textit{Algorithmic evolution brings benefits.}}}
From simple controller methods like PID to complex approaches like DT-based methods, we observe consistent performance improvements across all settings.
For instance, in Setting 1, BidFlow achieves a 95\% improvement in conversion while maintaining superior cost efficiency compared to PID. The sparse dataset shows similar trends with 3× better performance.
In Settings 2 and 3, advanced methods demonstrate superior adaptability. While PID and RL methods exhibit severe performance imbalance, sophisticated algorithms like DT score maintain balanced, high performance.
These empirical results demonstrate that algorithmic evolution yields compounding benefits, not only in raw conversion metrics but also in cost efficiency and robustness across diverse competitive environments.

{\textbf{\textit{Competitive and cooperative insights.}}}
The results in Table \ref{tab:dense_vs_sparse} demonstrates a strong negative correlation between CPA Ratio Variance and overall Score across both Dense and Sparse datasets. BidFlow, with the lowest variance, achieves the highest Scores, while PID, with the highest variance, shows the poorest performance.
This pattern reveals that low-variance strategies exhibit cooperative characteristics rather than aggressive competition, enabling multiple advertisers to achieve their targets simultaneously. Conversely, high-variance strategies reflect destructive over-competition, causing budget exhaustion and reduced platform revenue.
This confirms that variance serves as a reliable proxy for competition intensity, where stability promotes collective welfare.

\vspace{-3mm}
\subsection{Online Experiment Validation}
To verify the effectiveness of BidFlow in practice, we deployed it on the Kuaishou live streaming ad platform. Under this platform, advertisers set budgets with CPA/ROI constraints, and we compared BidFlow against the in-production and heavily tuned DT method with dedicated reward design, which is the strongest baseline.
The online A/B test was conducted in two isolated environments with identical and fair traffic allocations (each has 33\% of the whole traffic). 
This configuration corresponds to Setting 2, and Table \ref{tab:online} presents the results based on a 13-day online A/B test.

The significant improvements in platform revenue (Cost) and conversions (Target Cost) not only demonstrate BidFlow's superiority but also validate the benchmark's practical value for maintaining offline-online consistency. 
Additionally, \textbf{BidFlow has been fully deployed in Kuaishou live streaming ad platform}.
\vspace{-3mm}
\begin{table}[h!]
\centering
\caption{Online A/B test results on Kuaishou live streaming ad platform. Performance improvement of BidFlow compared to DT.}
\vspace{-3mm}
\label{tab:online}
\resizebox{0.48\textwidth}{!}{
\begin{tabular}{l|ccc}
\hline
                 & Impression $\uparrow$ & Cost $\uparrow$ & Target Cost $\uparrow$ \\ \hline
\textit{improve}      & +0.27\%   &+0.30\%     & +0.68\%     \\ \hline
\end{tabular}
}
\end{table}

\vspace{-2mm}
\subsection{Cross-Dataset Validation on iPinYou}
To verify that PlatformBid generalizes beyond AuctionNet, we instantiate the benchmark on the iPinYou dataset~\cite{iPinYou}, which differs structurally from AuctionNet in scale, advertiser pool, and auction characteristics. We adapt all three settings to the iPinYou multi-advertiser pool while keeping the pipeline unchanged, and report the average number of clicks per advertiser (higher is better), following standard iPinYou practice. We compare BidFlow with two representative generative baselines, DT score and CBD.

\begin{table}[h]
\centering
\caption{Cross-dataset validation on iPinYou (average clicks per advertiser, higher is better).}
\vspace{-3mm}
\label{tab:ipinyou}
\small
\setlength{\tabcolsep}{3mm}
\begin{tabular}{@{}llc@{}}
\toprule
\textbf{Setting} & \textbf{Method} & \textbf{Avg Clicks $\uparrow$} \\
\midrule
\multirow{3}{*}{S1 (Homogeneous)}   & DT score & 309.7 \\
                                    & CBD      & 302.5 \\
                                    & BidFlow (ours) & \textbf{324.1} \\
\midrule
\multirow{3}{*}{S2 (Heterogeneous)} & DT score & 299.5 \\
                                    & CBD      & 291.6 \\
                                    & BidFlow (ours) & \textbf{306.2} \\
\midrule
\multirow{3}{*}{S3 (Promotional)}   & DT score & 316.5 \\
                                    & CBD      & 308.9 \\
                                    & BidFlow (ours) & \textbf{331.8} \\
\bottomrule
\end{tabular}
\vspace{-3mm}
\end{table}

As shown in Table~\ref{tab:ipinyou}, BidFlow ranks first across all three settings on iPinYou, with the heterogeneous setting (Setting 2) confirming that improvements on target advertisers do not come at the cost of degrading baseline advertisers. Combined with AuctionNet results and the Kuaishou deployment, this provides cross-platform validation across three structurally distinct environments—e-commerce, display RTB, and short-video—confirming that neither PlatformBid nor BidFlow is tied to a specific data source. Additional public datasets will be incorporated as they become available.

It is also worth noting that AuctionNet currently remains the only publicly available large-scale dataset capturing realistic competitive bidding dynamics, and recent state-of-the-art auto-bidding works~\cite{li2024gas,gao2025gave,guo2024aigb,li2025cbd} are likewise evaluated on a single dataset for this reason. PlatformBid is designed to be dataset-agnostic and we will incorporate additional public datasets into the benchmark as they become available.

\vspace{-2mm}
\section{Limitations and Conclusion}
There are still several limitations of this research. First, we focus exclusively on auto-bidding algorithms, while auction mechanism design also significantly impacts bidding performance and warrants further investigation. Second, we do not explicitly model inter-advertiser relationships—a challenging but important research direction given that real-world platforms may host millions of advertisers with complex competitive dynamics.
Our benchmark could also be a testbed for these important research directions.

This paper addresses the gap between existing DSP-centric auto-bidding benchmarks and current unified ad platforms' requirements. We propose PlatformBid, the first platform-level auto-bidding benchmark that formulates various representative competitive settings and enables systematic evaluation of diverse methods. We further introduce BidFlow, which leverages expressive flow-matching policies to handle dynamic multi-advertiser competition. BidFlow achieves state-of-the-art performance in most PlatformBid settings, with online experiments confirming strong offline-online consistency, \textit{e.g.}, a +0.68\% improvement of target cost. With the continued development of unified ad platforms, this work establishes a critical foundation for advancing platform-centric auto-bidding research.

\vspace{-2mm}
\section*{Acknowledgements}
This research is supported by the Big Data Computing Center of Southeast University and Kuaishou Technology. We would also like to thank Pengfei Lyu for his valuable contribution to the online experiments.



\bibliographystyle{ACM-Reference-Format}
\balance
\bibliography{main}


\begin{center}
    \section*{Appendix}
\end{center}

\appendix

\section{Method Details}
\label{implementary details}

For the PID controller, we set $K_p=0.1$, $K_i=0.01$, and $K_d=0.03$ in all settings. 
For the DT reweight variant, the reweighting coefficient is set to $w=0.2$ across all settings.
For GAS, we employ three critic networks for both dense and sparse datasets across all settings.
For BidFlow, the BC coefficient $\alpha$ is set as follows: $\alpha=10$ for dense data in Settings 1 and 3, $\alpha=0.3$ for dense data in Setting 2, $\alpha=1$ for sparse data in Setting 2, and $\alpha=0.03$ for sparse data in Settings 1 and 3.


\begin{table*}[h]
\centering
\caption{Key metrics comparison under PlatformBid setting 2: Heterogeneous Competition. Each result pair (Target, Average) shows the average performance of only the target advertisers by the evaluated method and the average performance of all advertisers. Result pairs are bold if ``Average'' is the best.}
\label{supp:setting2}
\scriptsize
\renewcommand{\arraystretch}{1.0}
\resizebox{1.00\textwidth}{!}{
\setlength{\tabcolsep}{1mm}{
\begin{tabular}{@{}llccccccccc@{}}
\toprule
\textbf{Dataset} & \textbf{Metric} 
& \textbf{PID} & \textbf{IQL} & \textbf{BCQ} & \textbf{DT score} & \textbf{GAS} & \textbf{GAVE} 
& \textbf{CBD} & \textbf{BidFlow} \\
\midrule
\multirow{5}{*}{\textbf{Dense}} 
& Conversion $\uparrow$        & 170.54, 254.17 & 309.17, 331.55 & 231.04, 265.90 & 333.17, 345.44 & 340.38, 347.55 & 355.71, 330.07 & 313.29, 303.46 & \textbf{346.88, 347.96} \\
& Budget Util $\uparrow$   & 0.34, 0.65 & 0.88, 0.93 & 0.85, 0.90 & 0.85, 0.89 & 0.89, 0.93 & \textbf{0.99, 0.97}  & 0.98, 0.94 & 0.93, 0.95 \\
& CPA Ratio $\downarrow$   & 2.78, 1.91 & 0.99, 1.03 & 1.19, 1.19 & 0.85, 0.96 & 0.91, 0.97 & 0.99, 1.06 & 1.20, 1.13 & \textbf{0.93, 0.93} \\
& Qualified Rate $\uparrow$ & 0.17, 0.40 & 0.83, 0.69 & 0.54, 0.52 & 0.63, 0.63 & \textbf{0.67, 0.67} & 0.46, 0.48 & 0.50, 0.57 & 0.50, 0.57 \\
& Score $\uparrow$         & 143.21, 216.74 & 285.24, 293.86 & 165.21, 190.03 & \textbf{321.96, 321.92} & 315.81, 315.93 & 287.78, 262.78 & 229.86, 233.34 & 312.69, 308.83 \\
\midrule
\multirow{5}{*}{\textbf{Sparse}} 
& Conversion $\uparrow$        & 11.88, 21.65 & 38.46, 32.59 & 37.92, 34.42 & \textbf{44.38, 37.50} & 36.96, 33.86 & 29.30, 32.32 & 42.71, 31.84 & 38.08, 28.77 \\
& Budget Util $\uparrow$   & 0.51, 0.49 & 0.99, 0.69 & 0.68, 0.66 & \textbf{0.77, 0.69} & 0.75, 0.68 & 1.00, 0.70 & 0.98, 0.67 & 0.86, 0.66 \\
& CPA Ratio $\downarrow$   & 2.08, 1.28 & 0.94, 0.77 & 0.63, 0.69 & \textbf{0.63, 0.65} & 0.72, 0.70 & 1.26, 0.84 & 0.89, 0.77 & 0.83, 1.13 \\
& Qualified Rate $\uparrow$ & 0.25, 0.15 & 0.42, 0.25 & 0.25, 0.29 & 0.13, 0.13 & \textbf{0.33, 0.33} & 0.33, 0.19 & 0.17, 0.19 & 0.33, 0.29 \\
& Score $\uparrow$         & 9.52, 20.47 & 34.42, 30.31 & 37.78, 33.59 & \textbf{43.91, 36.26} & 36.47, 32.77 & 21.20, 28.29 & 38.71, 29.60 & 35.97, 26.51 \\
\bottomrule
\end{tabular}
}
}
\end{table*}

\begin{table*}[h]
\centering
\caption{Key metric comparison under PlatformBid Setting 3: Promotional Competition. ``Pro'' denotes promotional advertisers, ``Non'' denotes non-promotional advertisers, and ``All'' denotes the average performance across all advertisers.}
\label{supp:setting3}
\scriptsize
\renewcommand{\arraystretch}{1.0}
\resizebox{1.00\textwidth}{!}{
\setlength{\tabcolsep}{3mm}{
\begin{tabular}{@{}lll*{9}{c}@{}}
\toprule
\textbf{Dataset} & \textbf{Type} & \textbf{Metric}
& \textbf{PID} & \textbf{IQL} & \textbf{BCQ} & \textbf{DT} & \textbf{DT score} & \textbf{GAS} & \textbf{GAVE}
& \textbf{CBD} & \textbf{BidFlow} \\
\midrule

\multirow{12}{*}{\textbf{Dense}}
& \multirow{4}{*}{Pro}
&Conversion $\uparrow$  & 550.71 & 674.00 & 543.14 & 673.14 & 682.57 & 690.43 & 657.71 & 456.86 & \textbf{727.29} \\
& & CPA Ratio$\downarrow$ & 1.49 & 1.10 & 1.15 & 1.06 & 0.98 & 0.95 & 1.15 & 1.32 & \textbf{0.91} \\
& & Exceed $\downarrow$ & 0.57 & 0.43 & 0.71 & 0.71 & 0.42 & \textbf{0.28} & 0.86 & 0.71 & 0.29 \\
& & Score $\uparrow$  & 519.80 & 587.70 & 429.46 & 555.49 & 634.28 & 661.49 & 456.13 & 347.37 & \textbf{701.48} \\
\cmidrule(lr){2-12}

& \multirow{4}{*}{Non}
& Conversion $\uparrow$  & 124.93 & 294.10 & 234.17 & 307.85 & 316.43 & \textbf{319.95} & 304.32 & 252.17 & 310.61 \\
& & CPA Ratio$\downarrow$ & 1.71 & 1.02 & 1.20 & 1.12 & 1.01 & 0.95 & 1.20 & 1.37 & \textbf{0.89} \\
& & Exceed $\downarrow$ & 0.68 & 0.49 & 0.85 & 0.61 & 0.51 & 0.41 & 0.66  & 0.71 & \textbf{0.34} \\
& & Score $\uparrow$  & 111.46 & 247.66 & 171.42 & 230.79 & 271.85 & 277.90 & 203.71 & 169.46 & \textbf{297.25} \\
\cmidrule(lr){2-12}

& \multirow{4}{*}{All}
& Conversion $\uparrow$  & 187.02 & 349.50 & 279.23 & 361.12 & 371.29 & \textbf{373.98} & 355.85 & 282.02 & 371.38 \\
& & CPA Ratio$\downarrow$ & 1.68 & 1.03 & 1.19 & 1.11 & 1.01 & 0.95 & 1.20 & 1.36 & \textbf{0.90} \\
& & Exceed $\downarrow$ & 0.67 & 0.48 & 0.83 & 0.63 & 0.49 & \textbf{0.40} & 0.69 & 0.71 & 0.69 \\
& & Score $\uparrow$  & 171.01 & 297.25 & 209.05 & 278.14 & 321.73 & 333.84 & 240.52 & 195.41 & \textbf{356.20} \\
\midrule

\multirow{12}{*}{\textbf{Sparse}}
& \multirow{4}{*}{Pro}
& Conversion $\uparrow$  & 24.86 & 71.00 & 54.57 & 26.00 & 64.71 & \textbf{78.86} & 48.57 & 60.86 & 71.86 \\
& & CPA Ratio $\downarrow$ & \textbf{0.47} & 0.95 & 0.67 & 0.96 & 0.65 & 0.71 & 1.49 & 0.75 & 0.68 \\
& & Exceed $\downarrow$ & 0.38 & 0.14 & \textbf{0.00} & 0.29 & \textbf{0.00} & \textbf{0.00} & 0.86  & 0.14 & \textbf{0.00} \\
& & Score $\uparrow$  & 24.85 & 64.24 & 54.57 & 25.38 & 64.71 & \textbf{78.85} & 24.70 & 59.57 & 71.86 \\
\cmidrule(lr){2-12}

& \multirow{4}{*}{Non}
& Conversion $\uparrow$  & 5.95 & 27.44 & 17.49 & 15.05 & 23.90 & 23.93 & 22.93 & \textbf{29.27} & 28.80 \\
& & CPA Ratio $\downarrow$ & 1.37 & 0.98 & \textbf{0.81} & 1.12 & 0.86 & 1.14 & 1.64 & 0.99 & 0.92 \\
& & Exceed $\downarrow$ & 0.44 & 0.39 & \textbf{0.20}  & 0.37 & 0.32 & 0.41 & 0.78 & 0.44 & 0.29 \\
& & Score $\uparrow$  & 5.86 & 22.50 & 16.93 & 13.00 & 23.10 & 21.08 & 11.79 & 24.58 & \textbf{25.67} \\
\cmidrule(lr){2-12}

& \multirow{4}{*}{All}
& Conversion $\uparrow$  & 8.71 & 33.79 & 22.90 & 16.65 & 29.85 & 31.94 & 26.67 & 33.88 & \textbf{35.08} \\
& & CPA Ratio $\downarrow$ & 1.24 & 0.98 & \textbf{0.79} & 1.10 & 0.83 & 1.08 & 1.62 & 0.95 & 0.88 \\
& & Exceed $\downarrow$ & 0.38 & 0.35 & \textbf{0.17} & 0.35 & 0.27 & 0.35 & 0.79 & 0.40 & 0.25 \\
& & Score $\uparrow$  & 8.63 & 28.59 & 22.42 & 14.81 & 29.17 & 29.51 & 13.67 & 29.68 & \textbf{32.41} \\
\bottomrule
\end{tabular}
}
}
\end{table*}

\section{Additional Analysis on Setting 2 Sparse}
\label{app:setting2_sparse}

Under Setting 2 with sparse rewards, BidFlow underperforms DT score on the target group (35.97 vs.\ 43.91 in Table~\ref{supp:setting2}). We provide two complementary ablations to disentangle the underlying cause, and clarify how this case relates to deployment-time conditions.

\noindent\textbf{Context.} In production, the Kuaishou advertising environment contains both dense conversion types (e.g., app invocations, where a user jumps from Kuaishou to an external app and produces frequent reward signals) and sparse conversion types (e.g., lead submissions, which are rare events). Setting 2 Sparse isolates the purely sparse case as a stress test, which does not reflect the mixed conversion distribution encountered at deployment. Setting 2 Sparse should therefore be read in conjunction with the dense counterpart, where BidFlow is competitive.

\noindent\textbf{Ablation 1: Effect of inference architecture and Q-guidance.}
To test whether the one-step inference architecture limits BidFlow's expressiveness under sparse rewards, we compare three variants that share the same training data and BC flow backbone: the paper's one-step model with $\alpha=1$, a one-step variant with zero Q-guidance ($\alpha=0$), and the multi-step 50-step teacher used to produce distillation targets.

\begin{table}[h]
\centering
\caption{Ablation of inference architecture and Q-guidance on Setting 2 Sparse.}
\label{tab:ablation_alpha}
\small
\setlength{\tabcolsep}{2mm}
\begin{tabular}{@{}lccc@{}}
\toprule
\textbf{Variant} & \textbf{Score (Target)} & \textbf{Score (Avg)} & \textbf{CPA Ratio} \\
\midrule
1-step, $\alpha=1$ (ours) & 35.97 & 26.51 & 0.83 \\
1-step, $\alpha=0$         & 29.92 & 22.15 & 0.88 \\
50-step teacher            & 30.44 & 22.69 & 1.01 \\
\bottomrule
\end{tabular}
\end{table}

The 1-step $\alpha=0$ variant (29.92) and the 50-step teacher (30.44) yield nearly identical scores while differing only in inference depth. This rules out inference architecture as the primary cause of the gap to DT score. The lift from $\alpha=0$ to $\alpha=1$ (+6 points) further confirms that Q-value guidance contributes meaningfully, but does not fully close the gap, motivating a separate explanation.

\vspace{1mm}
\noindent\textbf{Ablation 2: Floor price sensitivity and collusive equilibrium.}
We note that DT score's CPA ratio in Setting 2 Sparse is only 0.63, well below the constraint, which indicates that it is not competing aggressively for impressions. Since DT score and the DT baseline share the same transformer backbone and differ only mildly in their reward design, they tend to produce correlated bidding behavior and form a low-price equilibrium that benefits DT-based strategies. This is consistent with the collusive bid suppression phenomenon documented in MAAB~\cite{multi-agentDB}, where agents with similar architectures trained on shared data develop correlated low-bid strategies that suppress market prices.

To counter collusive bid suppression, floor price mechanisms are standard practice~\cite{multi-agentDB}. In PlatformBid we apply a uniform floor price ratio (FPR) of $0.80\times$ the historical market price across all settings, analogous to MAAB-fix. To test whether DT score's advantage is sensitive to this floor, we increase FPR from $0.85$ to $0.90$ while holding all other components fixed.

\begin{table}[h]
\centering
\caption{Floor price sensitivity on Setting 2 Sparse (Heterogeneous). Score reported as Target / Avg.}
\label{tab:fpr_s2}
\small
\setlength{\tabcolsep}{2mm}
\begin{tabular}{@{}llccc@{}}
\toprule
\textbf{FPR} & \textbf{Method} & \textbf{Score (Target)} & \textbf{Score (Avg)} & \textbf{CPA Ratio} \\
\midrule
0.85 & DT score & 25.79 & 20.07 & 0.75 \\
0.85 & BidFlow  & \textbf{31.96} & \textbf{24.35} & 0.90 \\
\midrule
0.90 & DT score & 24.71 & 20.33 & 0.87 \\
0.90 & BidFlow  & \textbf{27.90} & \textbf{20.71} & 0.93 \\
\bottomrule
\end{tabular}
\end{table}

\begin{table}[h]
\centering
\caption{Floor price sensitivity on Setting 1 and Setting 3 Sparse.}
\label{tab:fpr_s1s3}
\small
\setlength{\tabcolsep}{2mm}
\begin{tabular}{@{}llcc@{}}
\toprule
\textbf{Setting} & \textbf{Method (FPR=0.85 / 0.90)} & \textbf{Score} & \textbf{CPA Ratio} \\
\midrule
\multirow{4}{*}{S1 (Homogeneous)}
 & DT score (0.85) & 25.75 & 0.92 \\
 & BidFlow  (0.85) & \textbf{28.57} & 0.88 \\
 & DT score (0.90) & 25.33 & 0.83 \\
 & BidFlow  (0.90) & \textbf{26.23} & 0.87 \\
\midrule
\multirow{4}{*}{S3 (Promotional)}
 & DT score (0.85) & 26.13 & --- \\
 & BidFlow  (0.85) & \textbf{30.78} & 0.86 \\
 & DT score (0.90) & 24.39 & 0.85 \\
 & BidFlow  (0.90) & \textbf{28.88} & 0.93 \\
\bottomrule
\end{tabular}
\end{table}

The results show a clear contrast between Setting 2 and the other two. In Setting 2, DT score's Avg Score degrades faster than BidFlow's as FPR is raised (DT score: $20.07 \rightarrow 20.33$; BidFlow: $24.35 \rightarrow 20.71$), and BidFlow leads at both FPR levels. In Settings 1 and 3, BidFlow leads DT score at every FPR value. This indicates that DT score's advantage under sparse rewards is exclusive to the heterogeneous setting, where correlated bidding with the DT baseline population produces a low-price equilibrium, rather than a general superiority over BidFlow. Adaptive floor price mechanisms such as MAAB's learned bar agent are orthogonal to PlatformBid's evaluation focus and constitute a natural direction for future work.

\section{Discussion on Data Realism}
\label{app:data_realism}

The realism of PlatformBid's underlying data is a natural concern for a benchmark intended to guide deployment decisions. PlatformBid is currently built on AuctionNet~\cite{su2024auctionnet}, which has been validated against real-world traffic from multiple complementary perspectives. At the feature level, the offline and online data distributions overlap substantially and exhibit the same cluster structure. At the value level, predicted pCVR distributions are consistent with those observed online. At the behavioral level, consumption distributions follow the same long-tail patterns as live traffic. AuctionNet is therefore a representative source for constructing platform-side auto-bidding benchmarks.

A second concern is whether external bidders with budgets and strategies outside the auto-bidding service affect the conclusions. On modern large-scale platforms, such bidders typically correspond to Real-Time API (RTA) advertisers, who interact with the platform via API on a per-impression basis. On platforms such as Kuaishou, the majority of advertisers adopt the auto-bidding service due to its effectiveness, and the RTA share remains small. This minority is unlikely to alter benchmark-level conclusions about auto-bidding methods.

Finally, the strongest evidence for real-world effectiveness is the online deployment in Section 5.3: BidFlow improves target cost by $+0.68\%$ on Kuaishou's production system, where the realized traffic distribution differs from the offline training data. This offline--online consistency supports the benchmark as a meaningful proxy for production decisions.

\end{document}